\RequirePackage{fix-cm}
\documentclass[a4paper]{article}
\usepackage[affil-it]{authblk}     
\usepackage{graphicx,subfigure}
\usepackage[bookmarks = true, colorlinks=true, linkcolor = black, citecolor = black, menucolor = black, urlcolor = black]{hyperref} 
\usepackage{amssymb}
\usepackage{color}
\usepackage{listings}
\usepackage{tikz}
\usepackage{verbatim}
\usetikzlibrary{arrows}
\usepackage{slashbox}
\begin{document}

\title{Neuron detection in stack images: a persistent homology interpretation\thanks{Partially supported by MEC project MTM2009-13842-C02-01, by the European Union's 7th 
Framework Programme under grant agreement nr. 243847 (ForMath), by MINECO project BFU2010-17537 and research founds from ONCE (years 2011--12).}}


\author{J\'onathan Heras, Gadea Mata, Germ\'an Cuesto, Julio Rubio, and Miguel Morales
}



\date{}

\maketitle

\begin{abstract}
Automation and reliability are the two main requirements when computers are applied in 
Life Sciences. In this paper we report on an application to \emph{neuron recognition}, an important
step in our long-term project of providing software systems to the study of neural morphology and
functionality from biomedical images. Our algorithms have been implemented in an ImageJ plugin
called NeuronPersistentJ, which has been validated experimentally. The \emph{soundness} and 
\emph{reliability} of our approach are based on the interpretation of our processing methods
with respect to \emph{persistent homology}, a well-known tool in computational mathematics.\\

\textbf{Keywords:} Neuron tracing; Dendrite recognition; Persistent Homology; Algebraic Topology.
\end{abstract}

\section{Introduction}\label{sec:intro}

The pioneer works of Ram\'on y Cajal suggested that \emph{neuronal morphology} and \emph{physiology} were intrinsically
correlated. The specificity of connections and information flux, as Cajal proposed, were closely dependent of \emph{neuronal structure}~\cite{Cajal88}.

Neuronal reconstruction and recognition were, since Cajal's work, hindered by a similar problem: the discerning
of a single neuron over hundreds of millions. Several staining techniques are employed to identify a single neuron~\cite{RM70};
for instance, Golgi staining, iontophoretic intracellular injection~\cite{ED02,BBEY06} and Diolostic gun~\cite{HBVC12}.
The use of optical and confocal microscopy and digital reconstruction of neuronal morphology has become a powerful technique
for investigating the nervous system structure, providing us with a large scale collection of images.

On the other hand, dendritic neuronal trees and axonal growing are involved in \emph{neuronal computation} and
\emph{brain functions}. Dendritic growing and axonal pathfinding are modified during brain development~\cite{Land94,Tes96}
neuronal plasticity process~\cite{GIHT11} and neural disorders such autism~\cite{Cal12} or degenerative diseases
such Alzheimer; for instance, in this neurodegenerative process brains are characterized by the presence of
numerous atrophic neurons near the amyloidal plaques~\cite{VP04,Goe06}. Therefore, visualization and analysis of
neuronal morphology and structure is of a critical importance to elucidate physiological changes.

The majority of reconstruction available software are manual or semiautomatic (see~\cite{Neurontracing}),
in which axonal and dendritic process are drawing by hand and consequently are not suitable for the analysis of
large arrays of data sets. Subsequently the traces would transform into a geometrical format suitable for
quantitative analysis and computational modeling. Algorithmic automation of neuronal tracing promises to increase
the speed, accuracy, and reproducibility of morphological reconstructions. In this way, large scale analysis is
feasible and would allow a high throughput strategy for the study of nervous system morphology in pharmacology
or degenerative diseases~\cite{DA11}. The properties of optical microscopes images make it
difficult to identify and automatically trace dendrites accurately, the presence of noise and biological
contaminations, i.e. dendritic segments from neighbors neurons make difficult the digital encoding and
reconstruction of a single neuronal structure.

To solve these problems, here, we employ geometric persistence models to extract the dendrites and neuronal
morphology from a series of inmunohistochemical images. The application developed is based on the idea
that the neuron that we want to study \emph{persists} in all the levels of the z-stack. The method presented
in this paper is not just theoretical but also has been implemented as a new plugin, called
\emph{NeuronPersistentJ}~\cite{NeuronPersistentJ}, for the systems ImageJ~\cite{ImageJ} and Fiji~\cite{Fiji}.

The rest of this paper is organized as follows. The following section is devoted to describe the cell cultures
and image acquisition methods. The procedure that we have developed for tracking neuronal morphology and
its interpretation in terms of persistent homology is presented in Section~\ref{sec:abam}. The experimental
results obtained with our software are discussed in Section~\ref{sec:er}. The paper ends with a discussion
section and the bibliography.

\section{Methods}\label{sec:bpaia}

\subsection{Cell cultures}

Primary cultures of hippocampal neurons were prepared from postnatal (P0-P1) rat pups as described previously in~\cite{Ban98,Mor00,C11}.
Briefly, glass coverslips ($12$ mm diameter) were coated with poly-L-lysine ($100$ $\mu$g/ml) and laminin ($4$ $\mu$g/ml). 
In brief, hippocampal neurons were seeded at 15000/cm$^2$ in culture medium consisting on Neurobasal medium (Invitrogen, USA)
supplemented with glutamine $0.5$ mM, $50$ mg/ml penicillin, $50$ units/ml streptomycin, $4\%$ FBS and $4\%$ B27 supplement 
(Invitrogen). At $4$, $7$, $14$ and $21$ days in culture, $100$ $\mu$l (from a $500$ ml total) of culture medium were replaced by $120$ ml
of fresh medium. At day $4$, $4$ $\mu$M of cytosine-D-arabinofuranoside was added to prevent overgrowth of glial cells.
Electroporation prior to plating was achieved using a square pulse electroporator (GenePulseXCell, BioRad). 
Usually, $4\times 106$ cells resuspended in $400$ $\mu$l of volumen, were mixed with $10$ $\mu$g plasmid (PDGF-GFP-Actin)
and electroporated following the following protocol: Voltage $200$ V, Capacitance $250$ $\mu$F in $400$ into a $4$ mm cuvette.

The expression vector encoding the protein chimeric with the N-terminus of chick b-actin under the control the platelet-derived
growth factor promoter region was kindly provided by Yukiko Goda~\cite{Col01}.

\subsection{Imaging and data analysis} 

Culture plates were mounted in the stage of a Leica DMITCS SL laser scanning confocal spectral microscope 
(Leica Microsystems Heidelberg, GmbH) with Argon laser attached to a Leica DMIRE2 inverted microscope.
For visualization and reconstruction of GFP-Actin, z-stacks images were acquired using $63$x oil immersion objective
lens (NA $1.32$), $488$ nm laser line, excitation beam splitter RSP $500$, emission range detection: $500-600$ nm, 
pixel size of: $58$ nm $\times$ $58$ nm and confocal pinhole set to 1 Airy units and $1,01$ $\mu$m between planes.
The maximum intensity projection is computed from the z-stack. 

\section{Algorithmic background and method}\label{sec:abam}

\subsection{The method}

\begin{figure}
\centering

\subfigure{%
            \label{maximum}
            \includegraphics[scale=0.1]{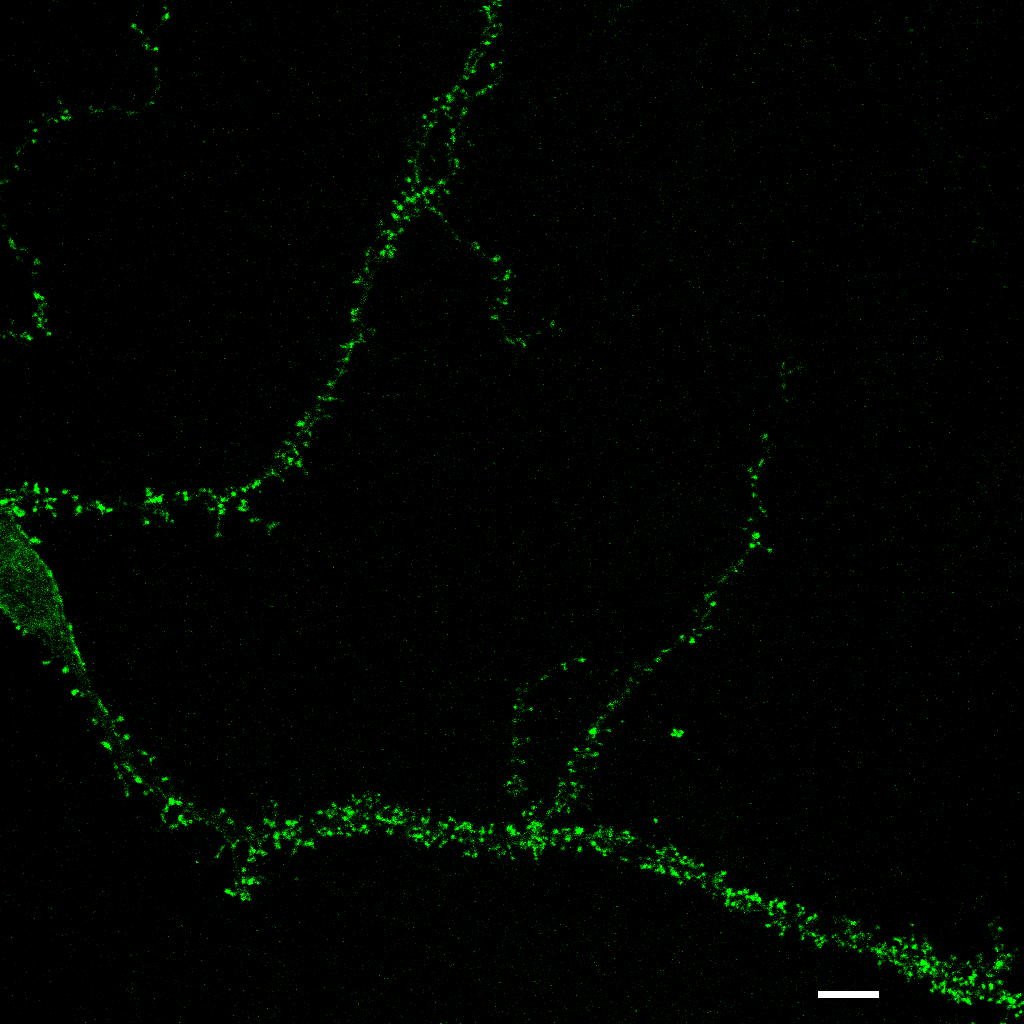}
        }%
        \subfigure{%
           \label{maximum_filter}
           \includegraphics[scale=0.1]{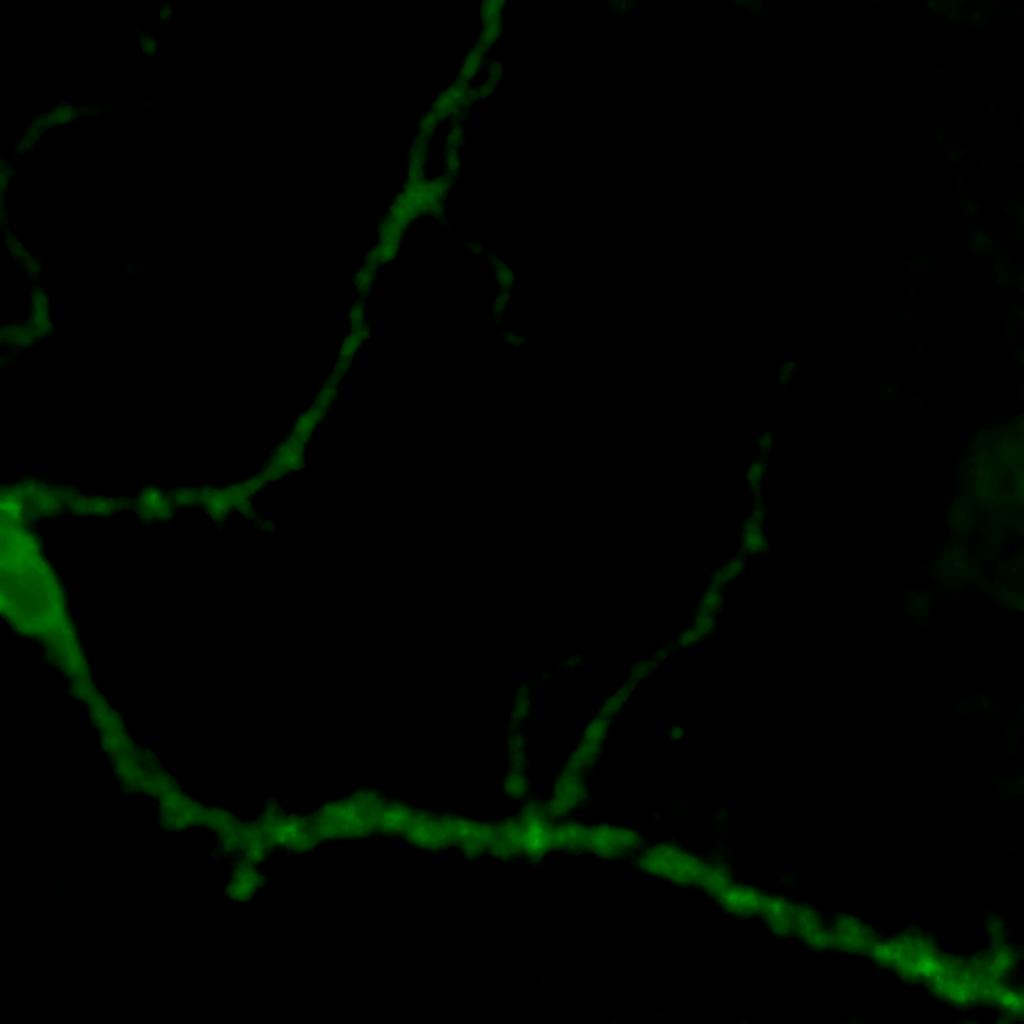}
        }
        \subfigure{%
            \label{maximum_filter_hold}
            \includegraphics[scale=0.1]{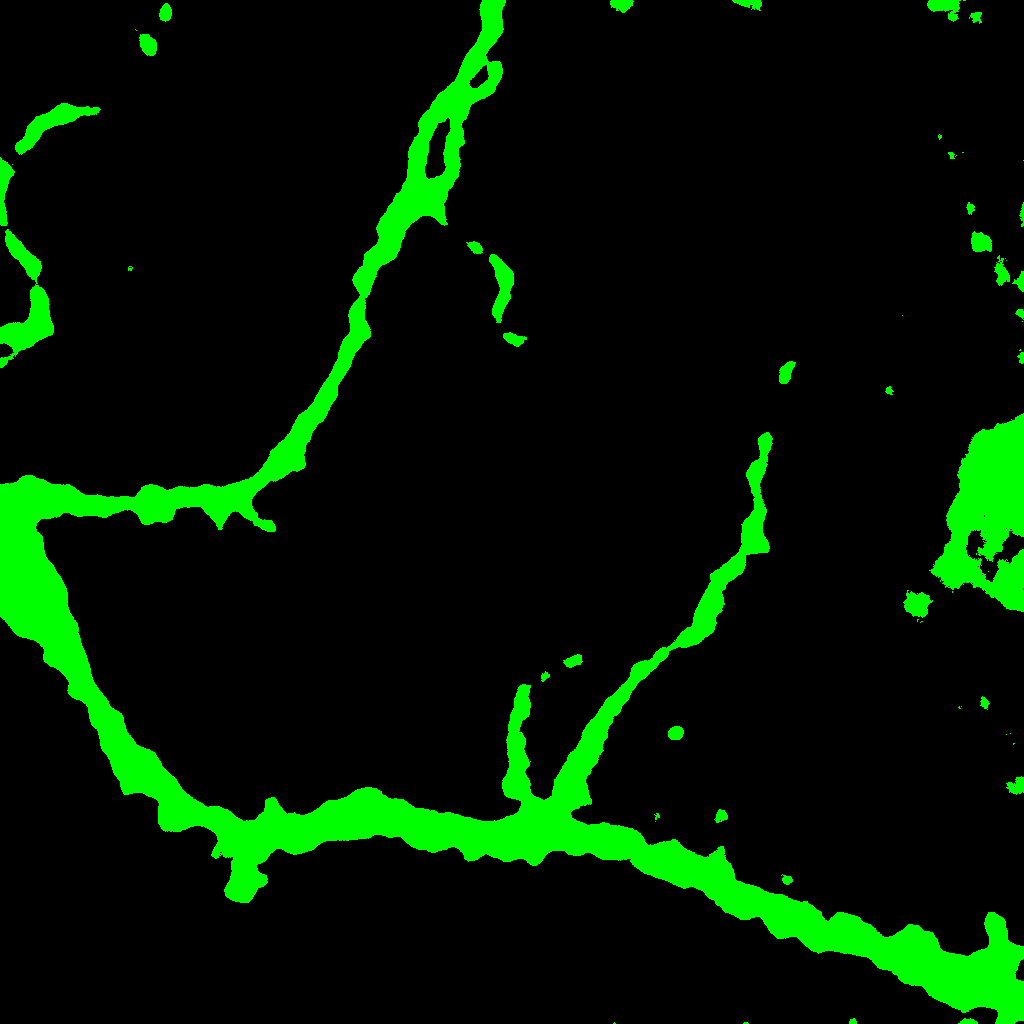}
        }
\caption{A 21 days in culture rat hippocampal neuron in culture, transfected with Actin-GFP. 
(a) Maximum intensity projection from a z-stack. (b) Median filter of the same image. (c) Huang's
thresholding method applied to the same image. Scale bar 5 $\mu$M.}\label{figure1}
\end{figure}

Our method to detect the neuronal structure from images, like the one of Figure~\ref{maximum}, can be split into two steps,
which will be called respectively \emph{salt-and-pepper removal} and \emph{persistent}. In the former one, we reduce the
salt-and-pepper noise, and in the latter one we dismiss the elements which appear in the image but which are not part of the
structure of the main neuron (astrocytes, other neurons and so on). 

In order to carry out the task of reducing the salt-and-pepper noise, we apply the following process both to the images
of the stack and to the maximum intensity projection image. Firstly, we apply a \emph{low-pass filter}~\cite{Cas96} to the images.
In our case, the filter which fits better with our problem is the \emph{median} one, since such a filter reduces speckle noise
while retaining sharp edges. The filter length is set to $10$ pixels for the situation described in Section~\ref{sec:bpaia}, 
this value has been pragmatically determined and it is the only parameter of the whole method which must be changed if
the acquisition procedure is modified. The result produced for the maximum intensity projection image of 
Figure~\ref{maximum} is shown in Figure~\ref{maximum_filter}.

Afterwards, we obtain binary images using \emph{Huang's method}~\cite{Huang}. This procedure automatically determines an adequate threshold 
value for the images. Applying that method to the image of Figure~\ref{maximum_filter}, we obtain the result depicted in 
Figure~\ref{maximum_filter_hold}.

However, in the image of Figure~\ref{maximum_filter_hold} we can see elements which does not belong
to the main neuronal structure. Let us explain how we manage to remove those undesirable elements. 

It is worth noting that in every slide of a z-stack appears part of the neuronal structure. On the contrary, 
irrelevant elements just appear in some of the slides. This will be the key idea of our method. 

\begin{figure}
\centering
\begin{tikzpicture}
\draw (0,0) node {\includegraphics[scale=0.075]{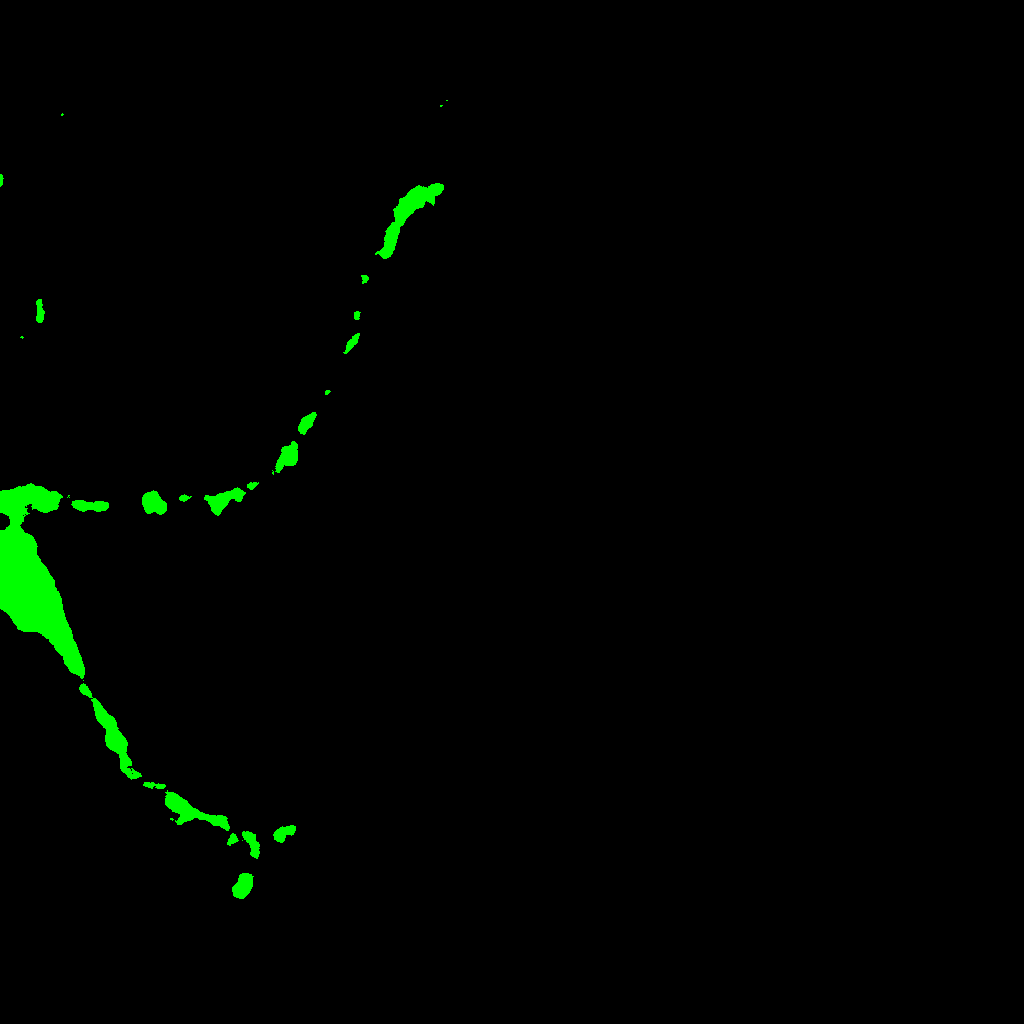}}; 
\draw (3,0) node {\includegraphics[scale=0.075]{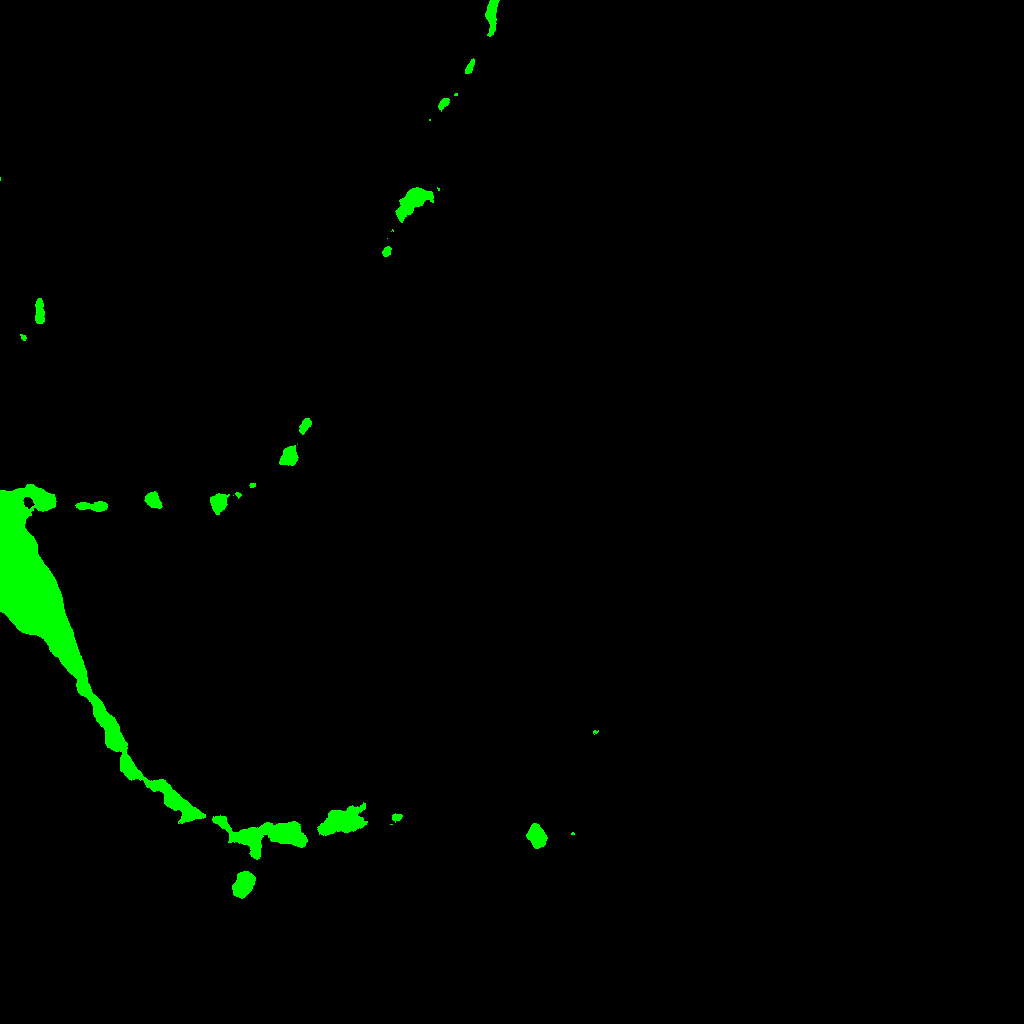}}; 
\draw (6,0) node {\includegraphics[scale=0.075]{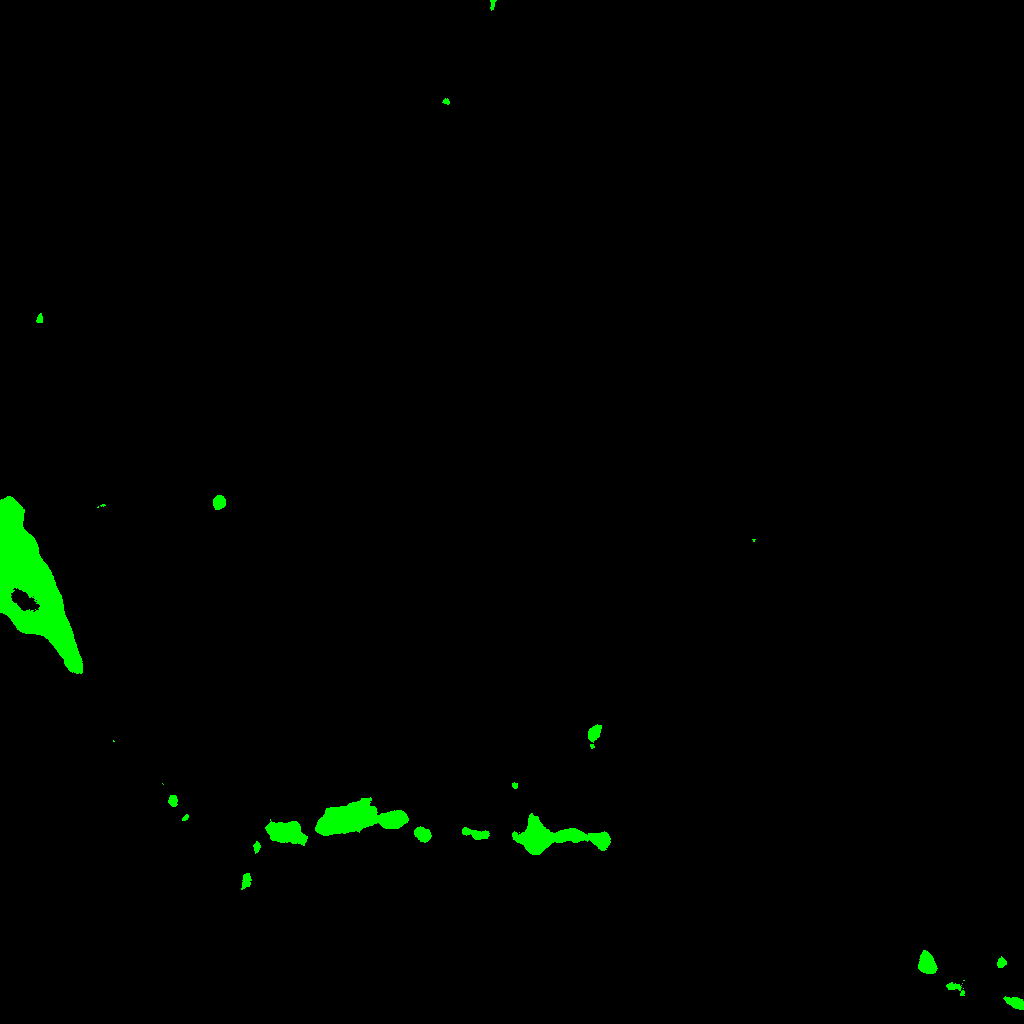}};
\draw (9,0) node {\includegraphics[scale=0.075]{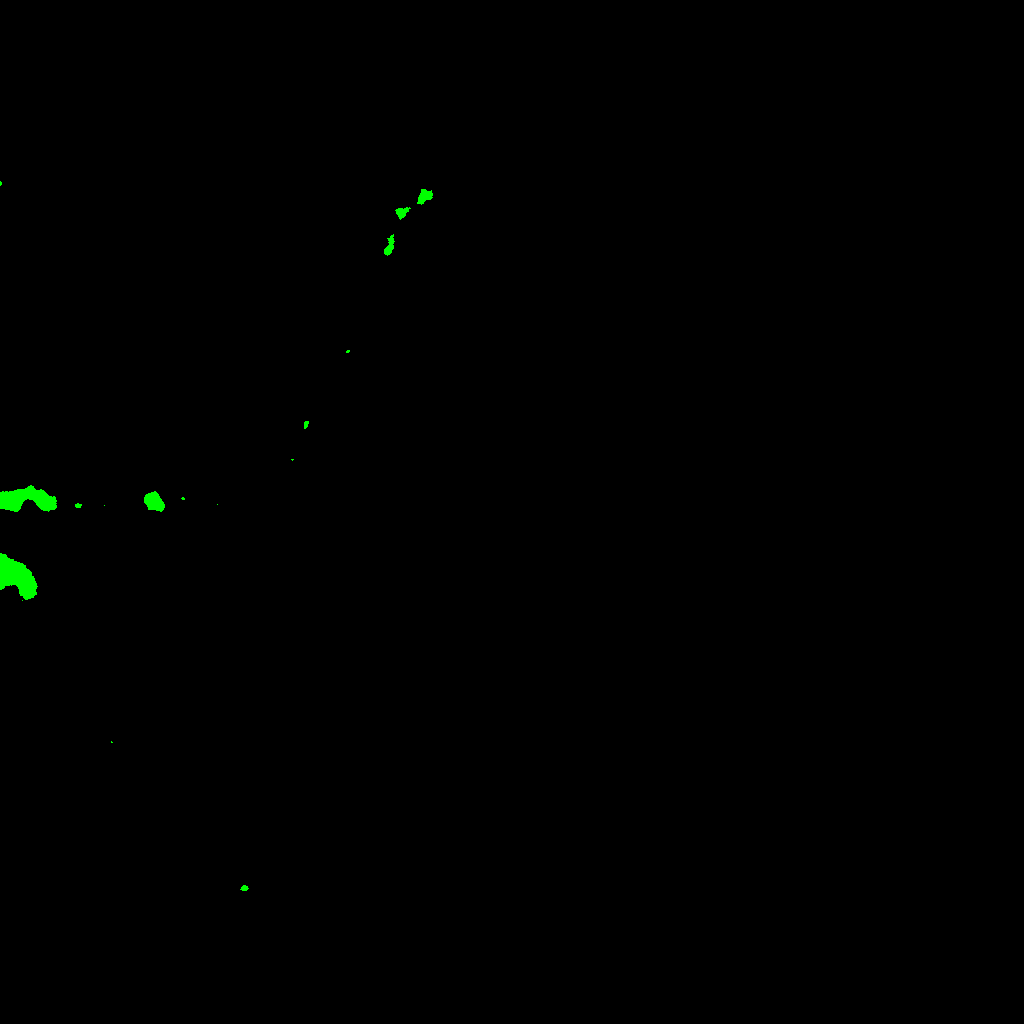}};  
\draw (0,-3) node {\includegraphics[scale=0.075]{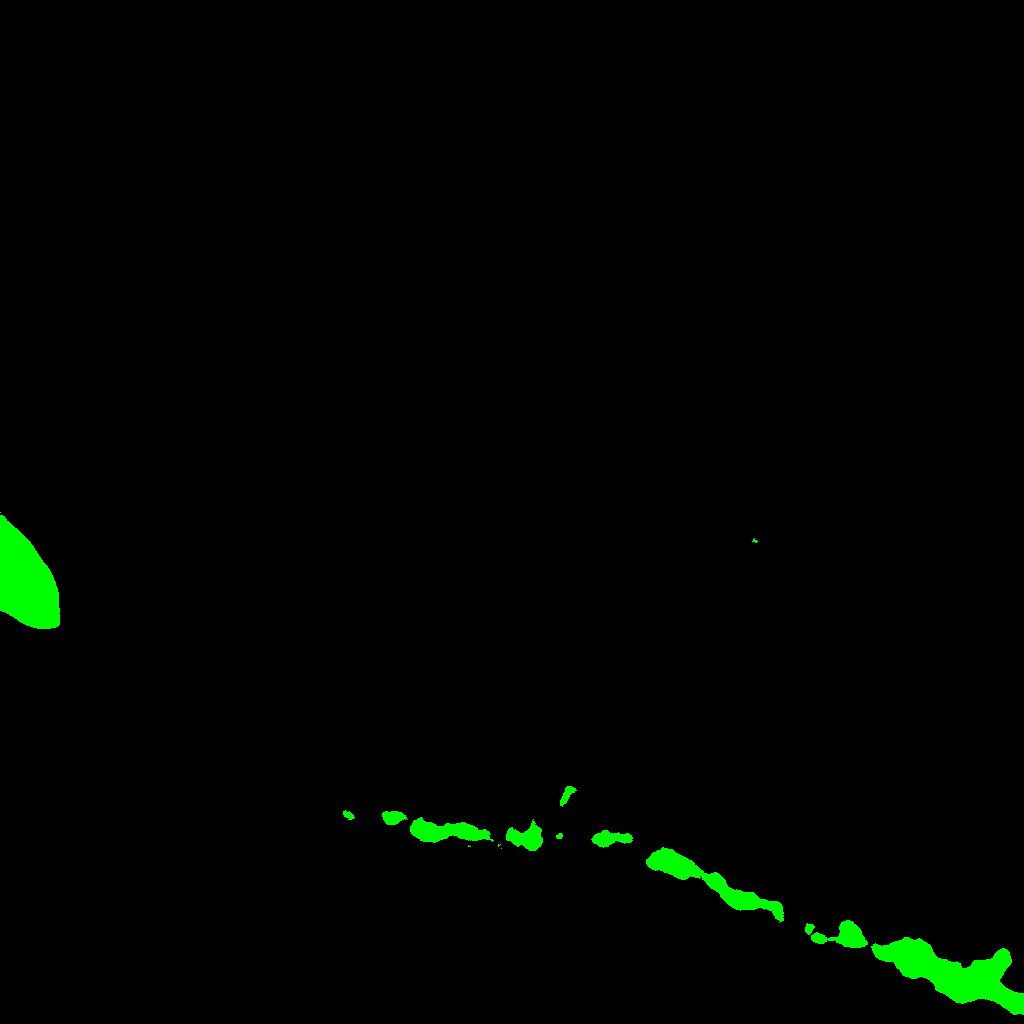}};  
\draw (3,-3) node {\includegraphics[scale=0.075]{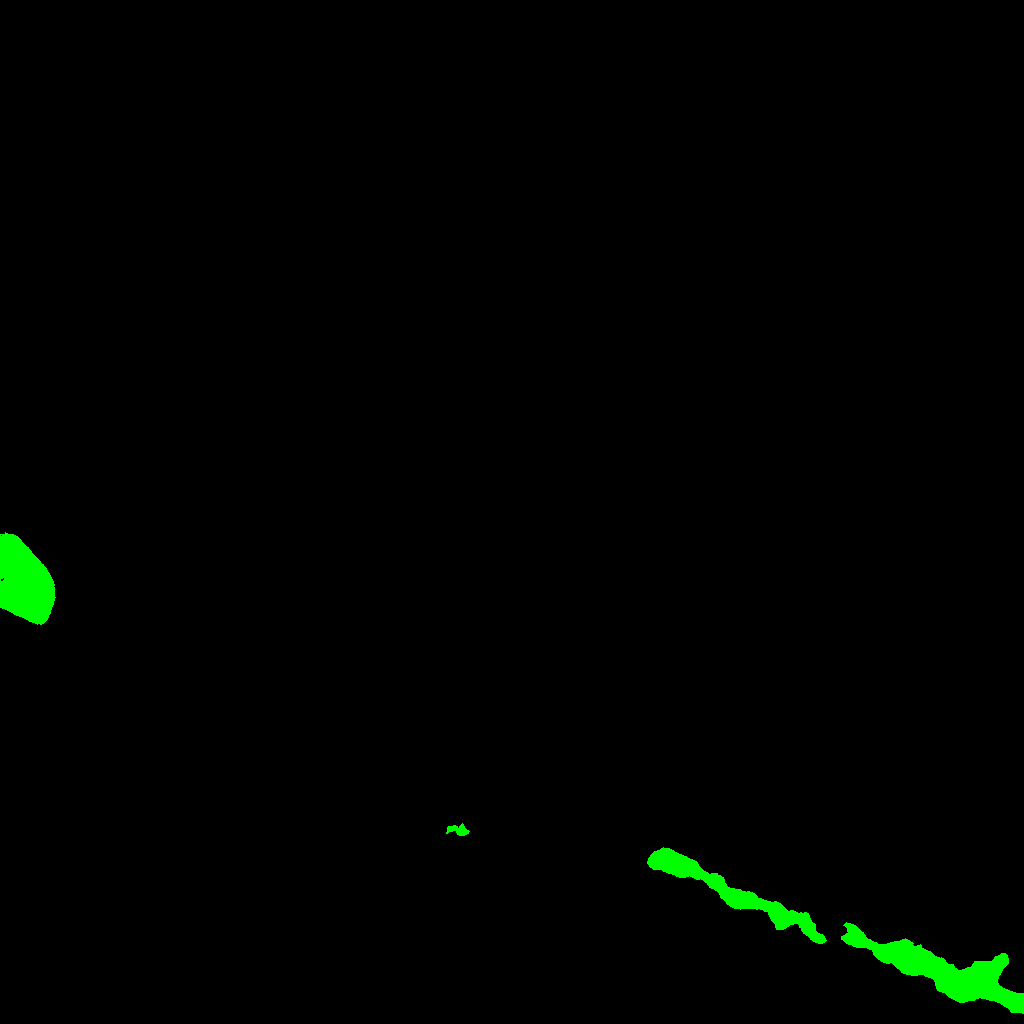}}; 
\draw (6,-3) node {\includegraphics[scale=0.075]{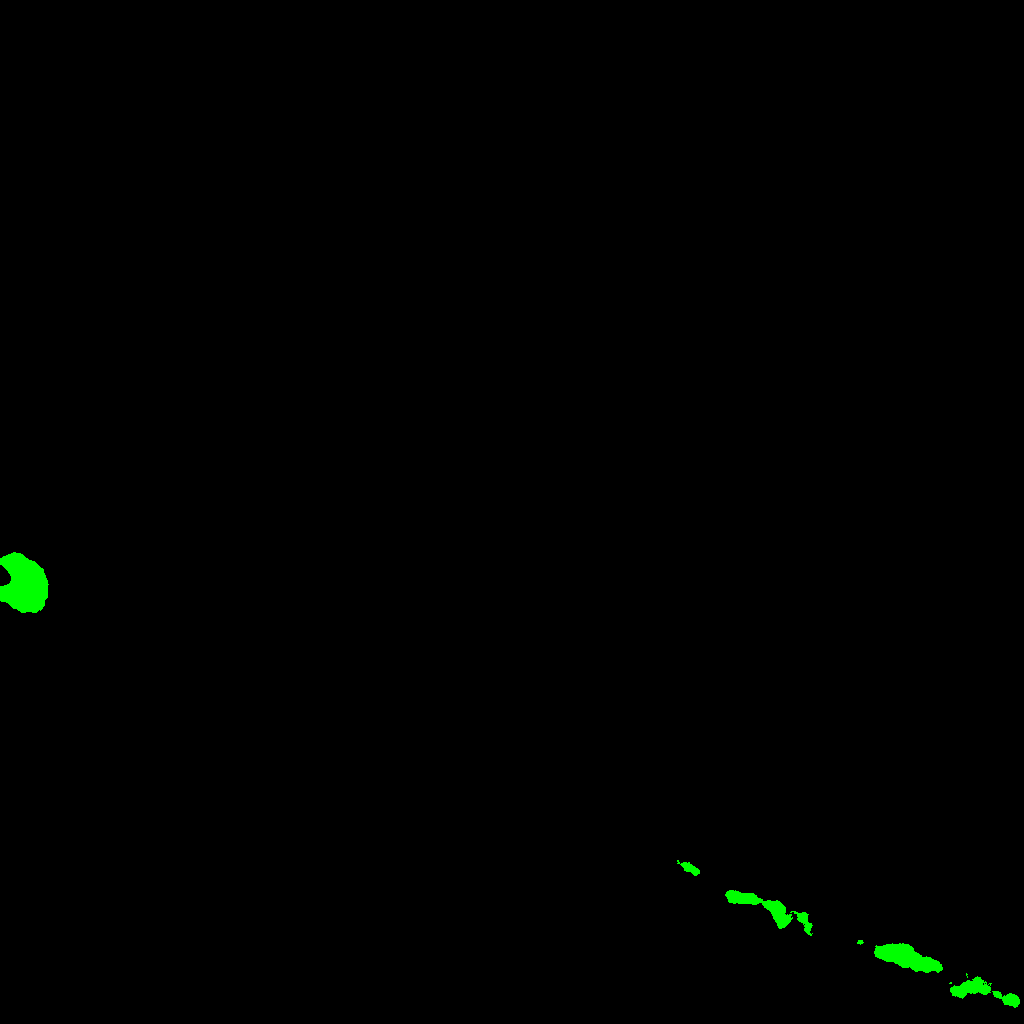}}; 
\draw (9,-3) node {\includegraphics[scale=0.075]{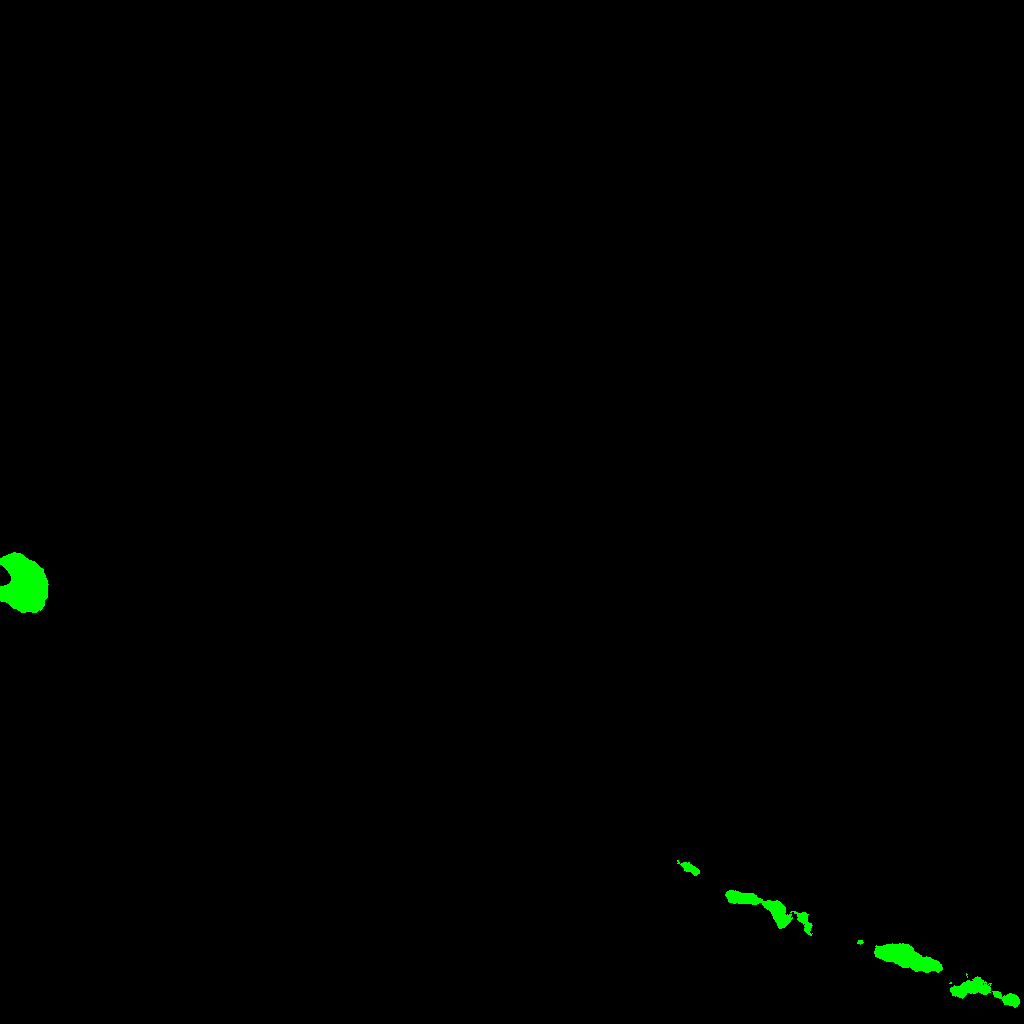}}; 
\end{tikzpicture}
\caption{Processed median and Huang's filter of each z-stack plane from Figure~\ref{figure1} neuron.}\label{processed_stack}
\end{figure}

More concretely, we proceed as follows. As we have explained previously, we apply the \emph{salt-and-pepper removal}
step to all the slides of the z-stack, the result of that in our case study can be seen in Figure~\ref{processed_stack}. 
 
In the \emph{persistent} step, we firstly construct a \emph{filtration} of the binary image associated with the maximum projection image. A
monochromatic image, ${\mathcal D}$, can be seen as a set of black pixels (which represent the foreground of the image), and
a \emph{filtration} of ${\mathcal D}$ is a nested subsequence of images $D^0 \subseteq D^1 \subseteq \ldots \subseteq D^m = \mathcal{D}$.
 
\begin{figure}
  \centering
  \begin{tikzpicture}
  \draw (0,0) node {\includegraphics[scale=0.06]{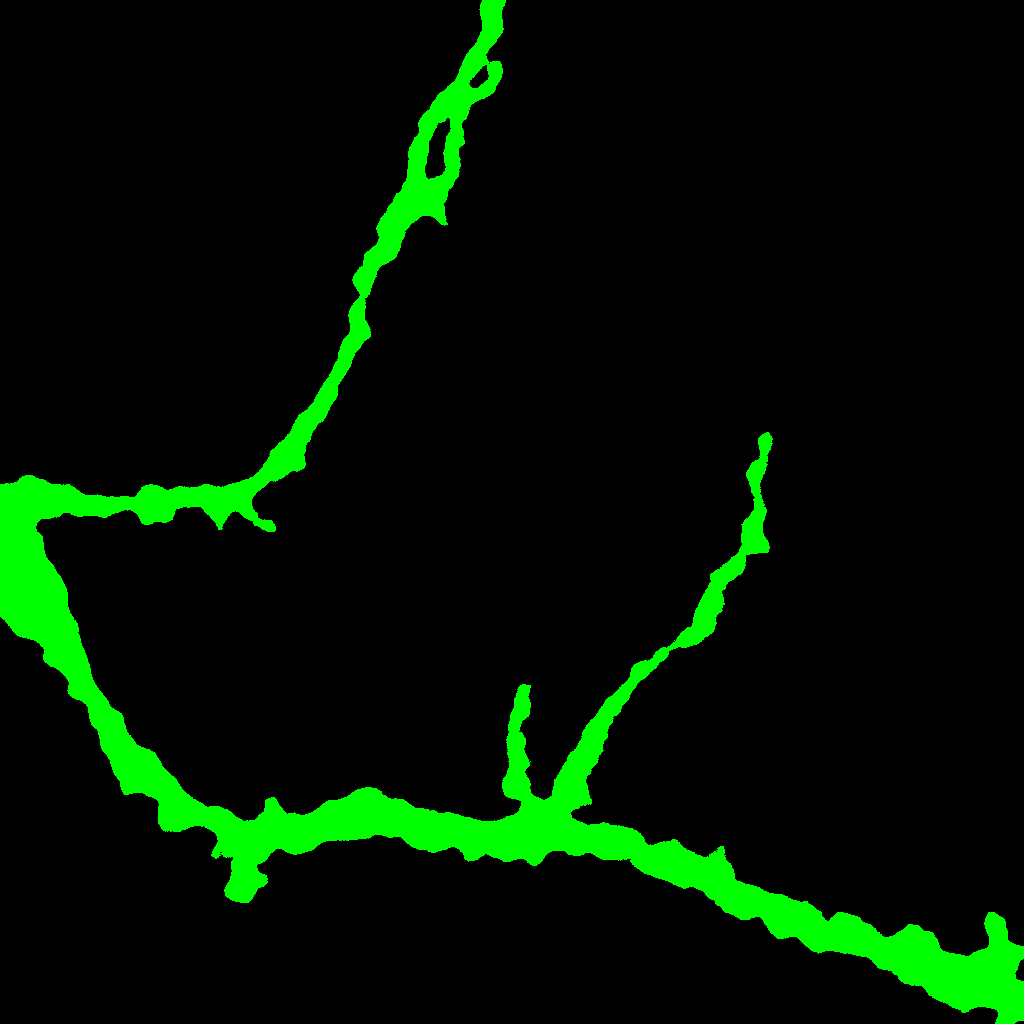}}; 
  \draw (1.5,0) node {$\subseteq$};
  \draw (0,-1.25) node {$D^0$};
  \draw (3,0) node {\includegraphics[scale=0.06]{5_8_hold_filt.png}}; 
  \draw (4.5,0) node {$\subseteq$};
  \draw (3,-1.25) node {$D^1$};
  \draw (6,0) node {\includegraphics[scale=0.06]{5_8_hold_filt.png}};
  \draw (7.5,0) node {$\subseteq$};
  \draw (6,-1.25) node {$D^2$};
  \draw (9,0) node {\includegraphics[scale=0.06]{5_8_hold_filt.png}};  
  \draw (10.5,0) node {$\subseteq$};
  \draw (9,-1.25) node {$D^3$};
  \draw (0,-3) node {\includegraphics[scale=0.06]{5_8_hold_filt.png}};  
  \draw (1.5,-3) node {$\subseteq$};
  \draw (0,-4.25) node {$D^4$};
  \draw (3,-3) node {\includegraphics[scale=0.06]{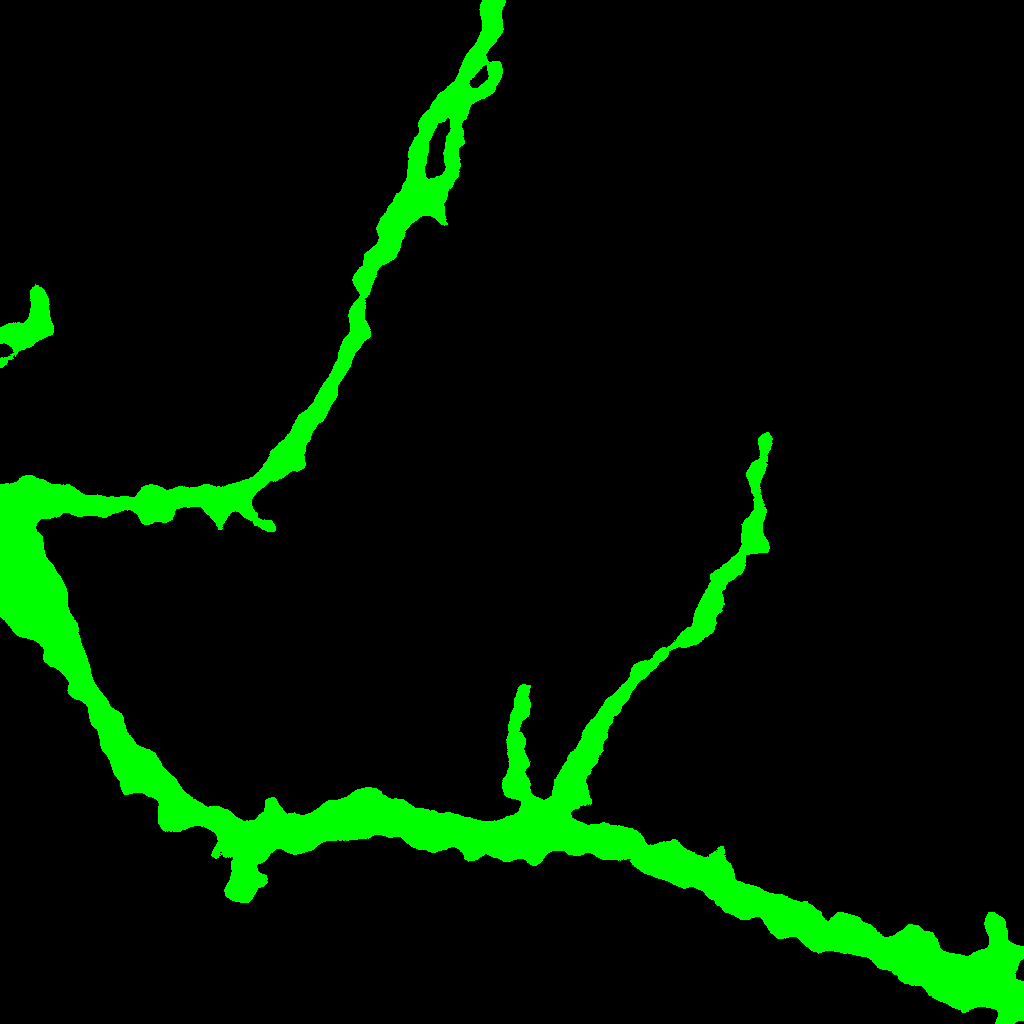}}; 
  \draw (4.5,-3) node {$\subseteq$};
  \draw (3,-4.25) node {$D^5$};
  \draw (6,-3) node {\includegraphics[scale=0.06]{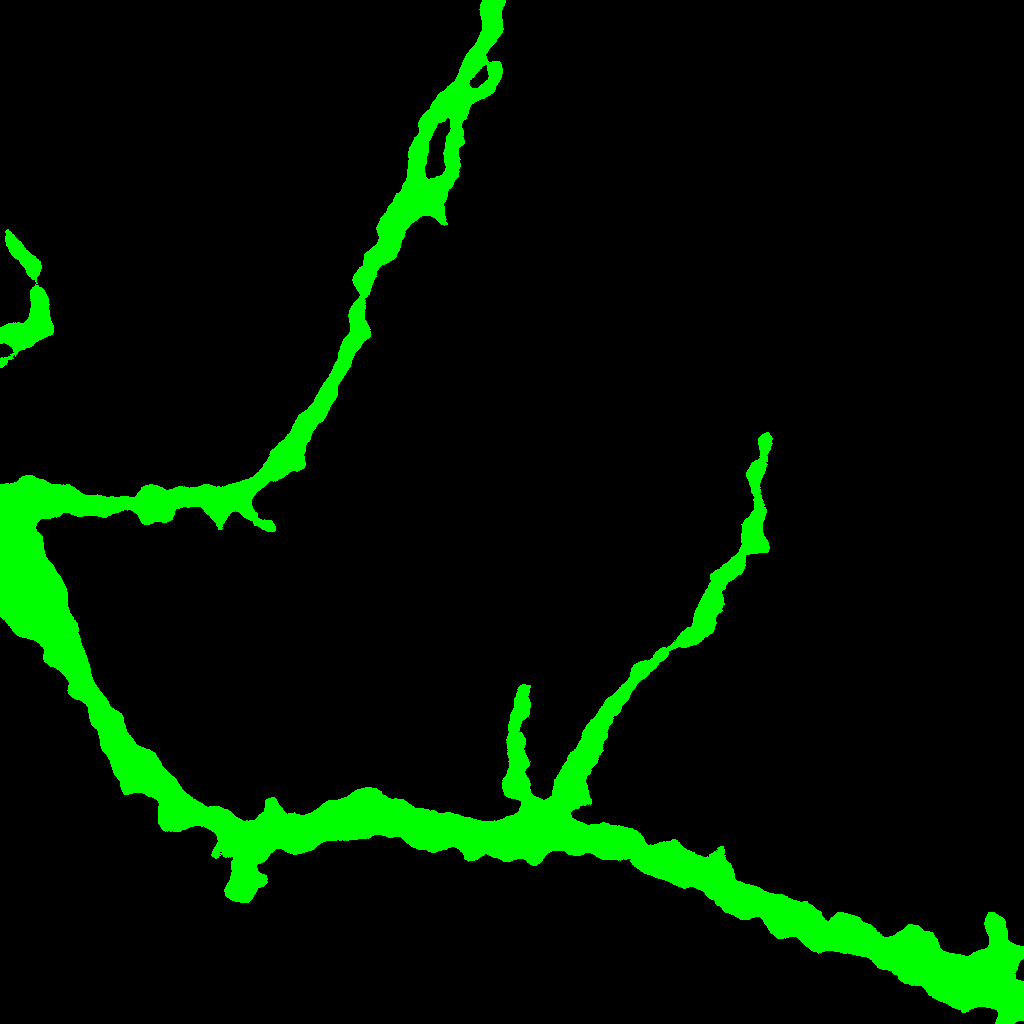}}; 
  \draw (7.5,-3) node {$\subseteq$};
  \draw (6,-4.25) node {$D^6$};
  \draw (9,-3) node {\includegraphics[scale=0.06]{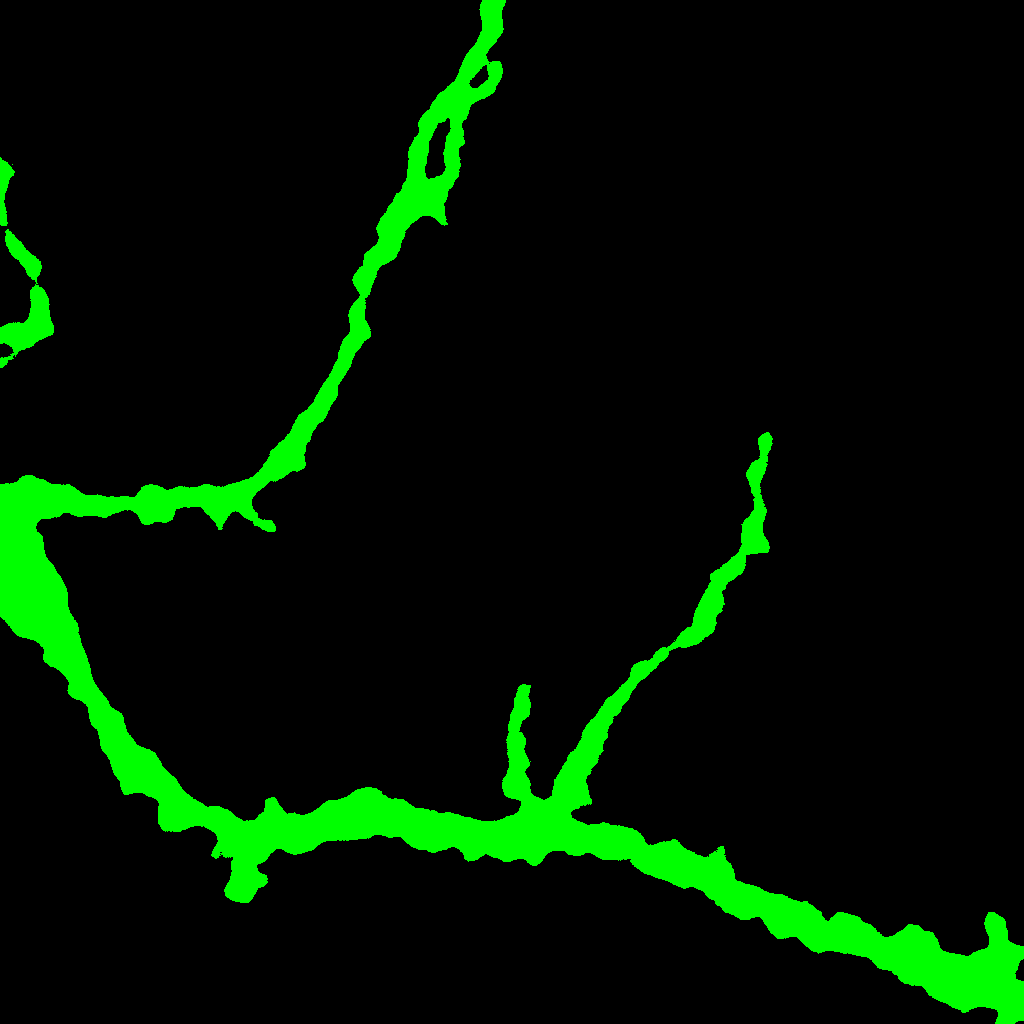}};
  \draw (10.5,-3) node {$\subseteq$};
  \draw (9,-4.25) node {$D^7$};
  \draw (4.5,-6) node {\includegraphics[scale=0.06]{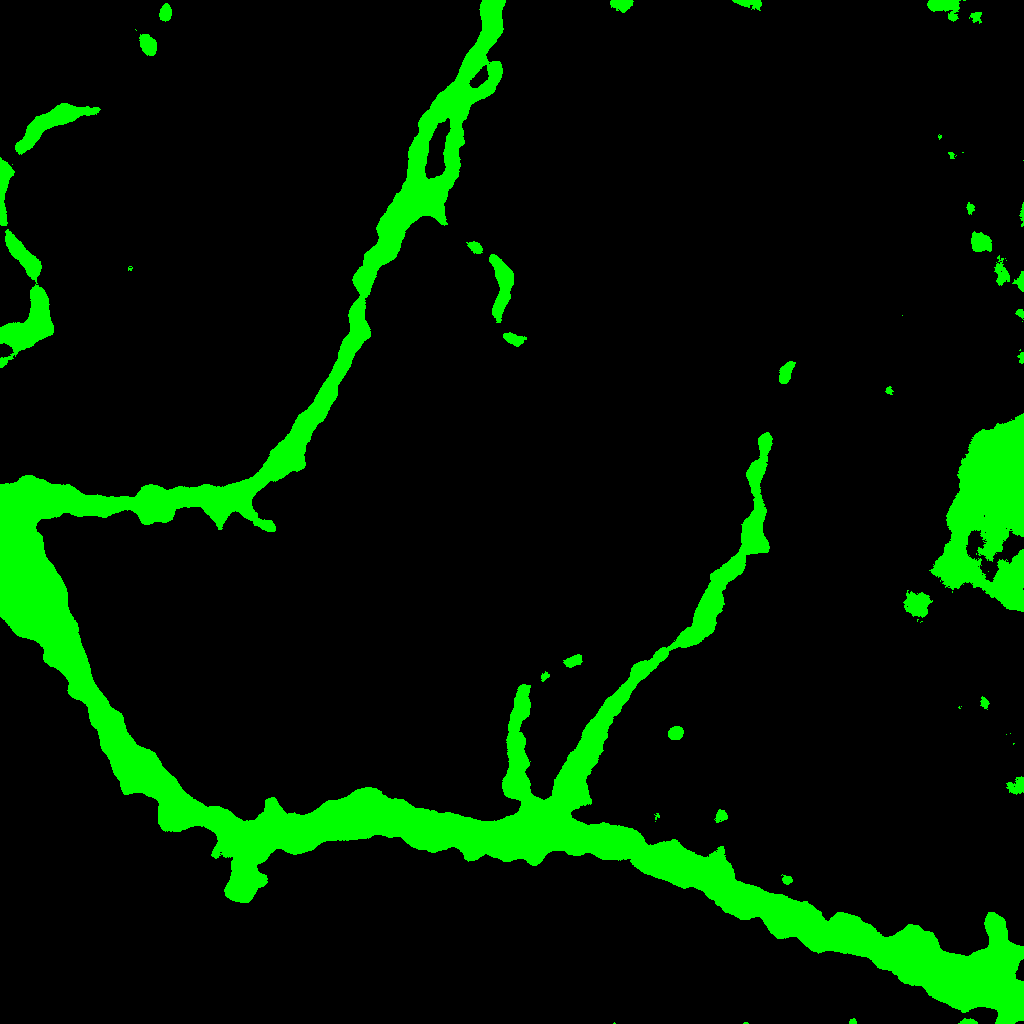}}; 
  \draw (4.5,-7.25) node {$D^8$};
  \end{tikzpicture}
  \caption{A series of pictures depicting the process of filtration from the z-stack of Figure~\ref{figure1}.
  From $D^0$ to $D^8$: Starting on $D^0$ and following to $D^8$ each level of the filtration represent the 
  containing, $\subseteq$ information from the previous level. $D^8$ contains all the connected components from 
  the image.}\label{filtration_processed_stack}
  \end{figure}

In order to construct a filtration of the binary image associated with the maximum projection image we proceed as follows. 
$D^m$ is the maximum projection image. $D^{m-1}$ consists of the connected components of $D^m$ 
whose intersection with the first slide of the stack is non empty. $D^{m-2}$ consists of the connected components of $D^{m-1}$ 
whose intersection with the second slide of the stack is non empty, and so on. In general,  $D^{m-n}$ consists of the connected components of $D^{m-n+1}$
whose intersection with the $n$-th slide of the stack is non empty. In this way, a filtration of the maximum projection image is obtained, see 
Figure~\ref{filtration_processed_stack}. 

As we know that the neuron appears in all the slides of the stack, the component $D^0$ of our filtration
will be the structure of the neuron. As a final remark, we can notice that the construction of the 
filtration reaches a point where it is stable; that is, a level of the filtration $D^i$ of the filtration 
such that $D^j$ is equal to $D^i$ for all $0\leq j < i$. An example can be seen in the components 
$D^0$ to $D^4$ of Figure~\ref{filtration_processed_stack}. This observation will be important in the next
subsection.

\subsection{Interpretation in terms of persistent homology}

The persistent adjective of the second step of the method presented in the previous subsection comes from
the nice interpretation which can be given in terms of the \emph{persistent homology theory}~\cite{ELZ02},
a branch of Algebraic Topology~\cite{Mau96}. In a nutshell, persistent homology is a technique which 
allows one to study the \emph{lifetimes} of \emph{topological attributes}. A detailed description of persistent homology can
be seen in~\cite{ELZ02,PHDZomorodian}, here we just present the main ideas.

One of the most important notions in Algebraic Topology is the one of \emph{homology groups}. The homology group in dimension
$n$ of an object $X$, denoted by $H_n(X)$, is a set which consists of the $n$ dimensional holes of $X$, also called \emph{$n$ dimensional
homology classes} of $X$. To be more concrete, $H_0(X)$ measures the number of connected components of $X$, and the
homology groups $H_n(X)$, with $n>0$, measure higher dimensional connectedness. In the case of $2$ dimensional monochromatic images, 
the $0$ and $1$ dimensional homology classes are, respectively, the connected components and the holes of the image; there are not homology classes 
in higher dimensions. 

Now, given a $2$ dimensional monochromatic digital image ${\mathcal{D}}$ and a filtration 
$D^0 \subseteq D^1 \subseteq \ldots \subseteq D^m = \mathcal{D}$ of ${\mathcal{D}}$, a $n$-homology class $\alpha$ \emph{is born} at $D^i$ if it belongs to the set $H_n(D^i)$ 
but not to $H_n(D^{i-1})$. 
Furthermore, if $\alpha$ is born at $D^i$ it \emph{dies entering} $D^j$, with $i<j$, if it belongs to the set $H_n(D^{j-1})$ but not to
$H_n(D^{j})$. The \emph{persistence} of $\alpha$ is $j-i$. We may represent the lifetime of a homology class as an interval, and we
define a \emph{barcode} to be the set of resulting intervals of a filtration. 

In the case of the filtrations presented in the previous subsection, the outstanding barcode is the one of $0$ dimensional homology classes.
It is worth noting that the structure of the neuron lives from the beginning to the end of the filtration while 
external elements are short-lived. 

\begin{figure}
\centering
\includegraphics[scale=.15]{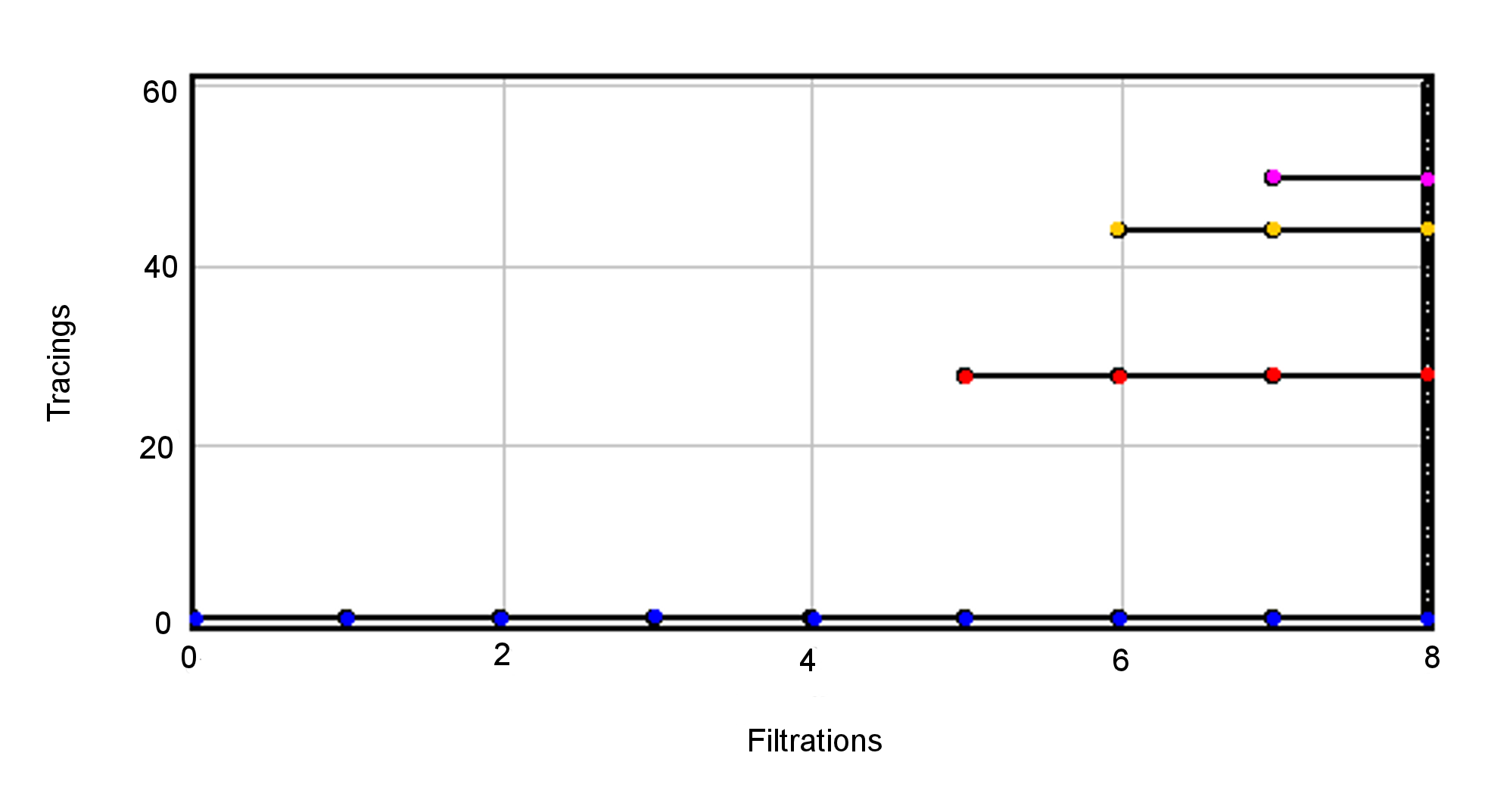}
\caption{Barcode summary of the filtration process from Figure~\ref{filtration_processed_stack} obtained using NeuronPersistentJ}\label{barcode}
\end{figure}

For example, the barcode associated with the filtration of Figure~\ref{filtration_processed_stack} is the one depicted in Figure~\ref{barcode}. 

\begin{figure}
\centering
\includegraphics[scale=0.2]{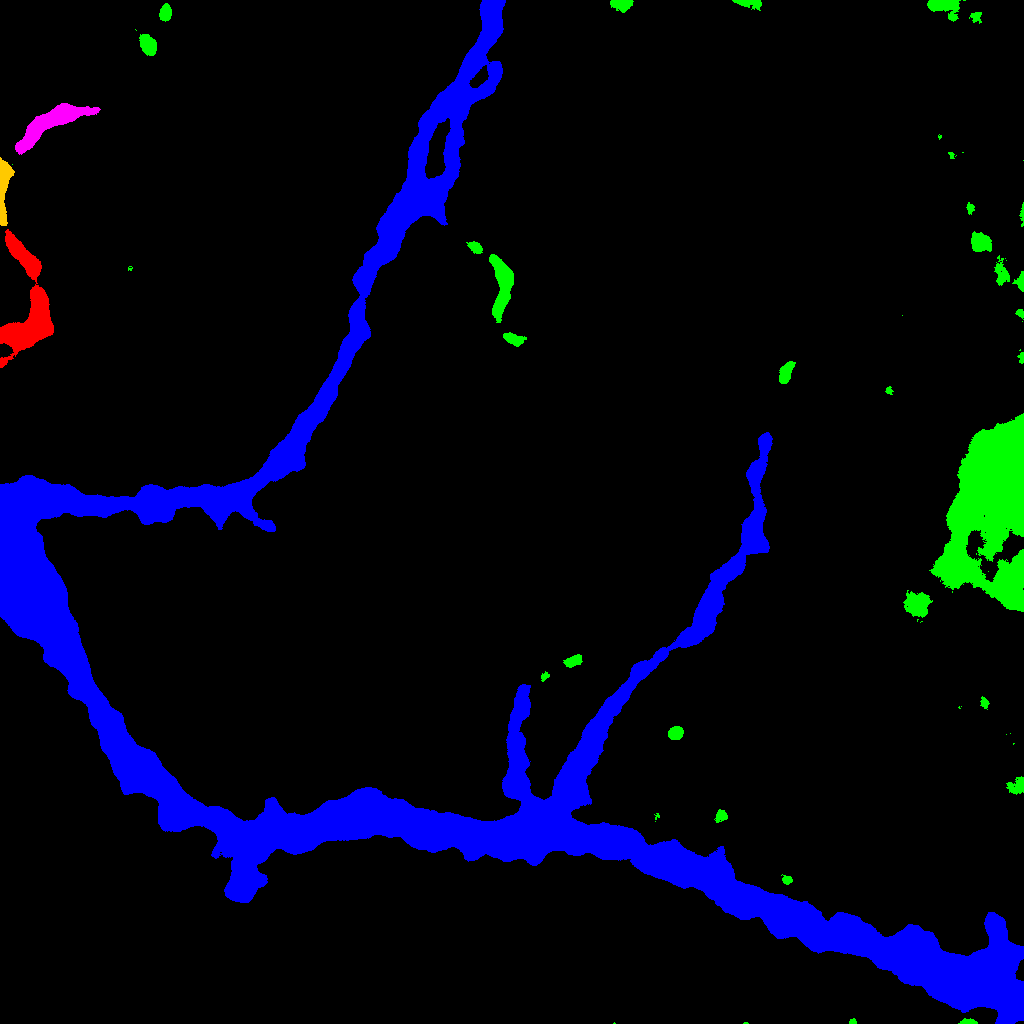}
\caption{Summary picture of the connected components from Figure~\ref{figure1} projection. 
 Color code: green, components that last one plane.
 Orange, 2 planes. Yellow, three planes. Red, four planes and blue, components that are present in the eight planes.
}\label{maximum_filter_hold_color}
\end{figure}

Let us analyze the information which can be extracted from such barcode. There are several connected components which live is reduced to the 
maximum projection image, the green connected components of Figure~\ref{maximum_filter_hold_color}, and can be considered as noise.
Notwithstanding that the components $x_1,x_2$ and $x_3$ (which are respectively the red, yellow and orange connected components of 
Figure~\ref{maximum_filter_hold_color}) live a bit longer than green components; they are also short-lived; so, they cannot be part of the main structure
of the neuron, it is likely that these components come from other biological elements. Eventually, we have the $x_0$ component, the blue
connected component of Figure~\ref{maximum_filter_hold_color}, which lives from the beginning to the end of the filtration; therefore, as it lives from the
beginning to the end of the filtration, it represents the structure of the neuron. 

We have devised an efficient algorithm to obtain the barcode of $0$ dimensional homology classes associated with
the images that we have presented in the previous subsection. This method takes advantage of both the way of building
the filtration and the stability of such a filtration. Firstly, we obtain the connected components of the level $0$
of the filtration, $D^0$; this is a well-known process called \emph{connected component labeling} which can be solved
using different efficient algorithms, see~\cite{ccl10,Lee07}. Such connected components are $0$ dimensional homology
classes which are born at $D^0$ and live until the end of the filtration, this fact comes from the filtration 
construction process. Now, we focus on the level $1$ of the filtration, $D^1$. As we have seen at the end of the
previous subsection, the filtration has a stability level; therefore, we consider two feasible cases. 
If $D^0$ is equal to $D^1$, we can pass to the next level of the filtration. Otherwise, we obtain the connected components
which appear at $D^1$ but not at $D^0$, such components are $0$ dimensional homology classes which are born at $D^1$
and live until the end of the filtration. In order to check if $D^0$ and $D^1$ are equal, we use the 
\emph{MD6 Message-Digest Algorithm}~\cite{MD6}. Such algorithm is a cryptographic hash function which given an image
returns a \emph{unique} string; therefore, if the result produced for $D^0$ and $D^1$ is the same, we can claim that both
images are equal. This procedure is faster than comparing pixel by pixel the images.

The above process is iterated for the rest of the levels of the filtration. In general, if we are in the level $i$ of
the filtration, there are two cases: if $D^{i-1} = D^i$ (this is tested with MD6 algorithm) then pass to level $i+1$; otherwise
the connected components which appear in $D^i$ but not in $D^{i-1}$ are the $0$ dimensional homology classes which are born at $D^i$
and live until the end of the filtration. In this way, we can obtain the barcode of $0$ dimensional homology classes without
explicitly computing persistent homology.

\section{Experimental results}\label{sec:er}

\begin{figure}
\centering

\subfigure{%
            \label{maximum1}
            \includegraphics[scale=0.1]{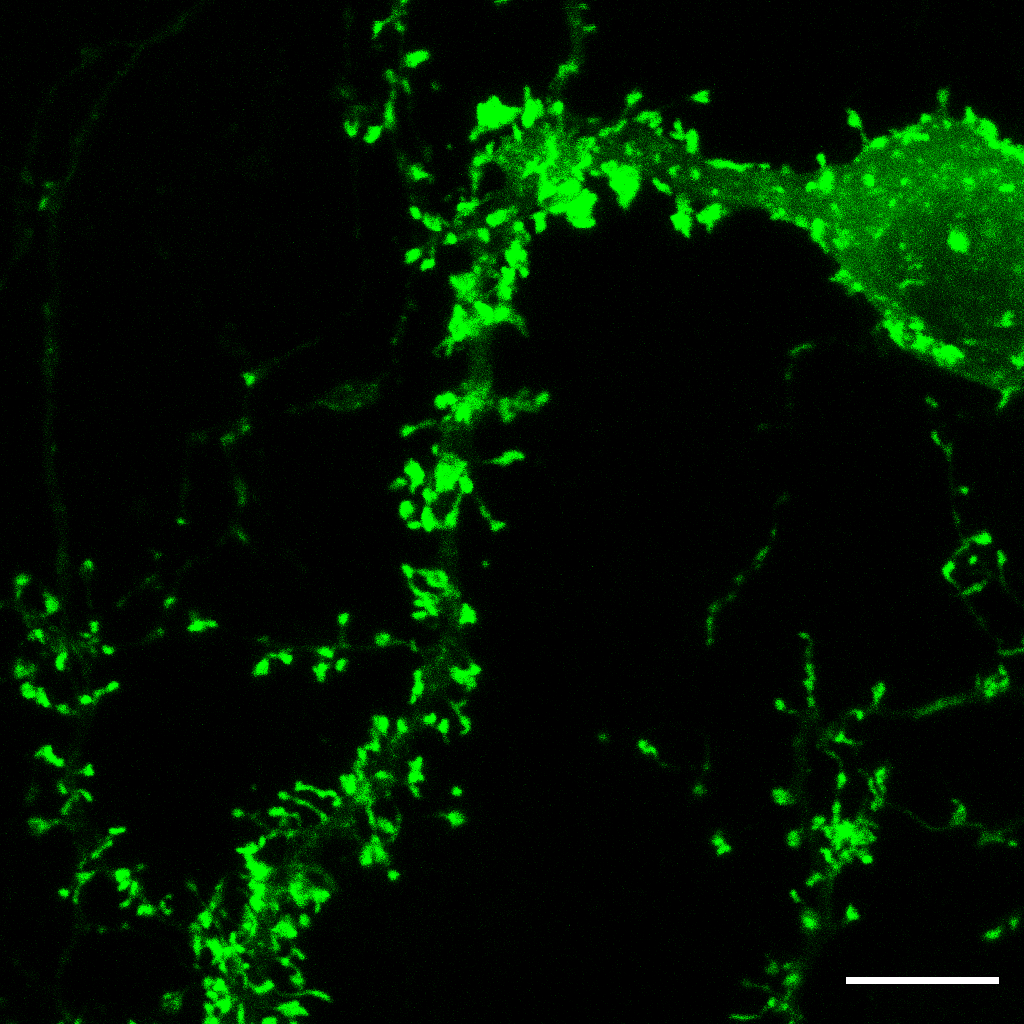}
        }%
        \subfigure{%
           \label{maximum_filter1}
           \includegraphics[scale=0.1]{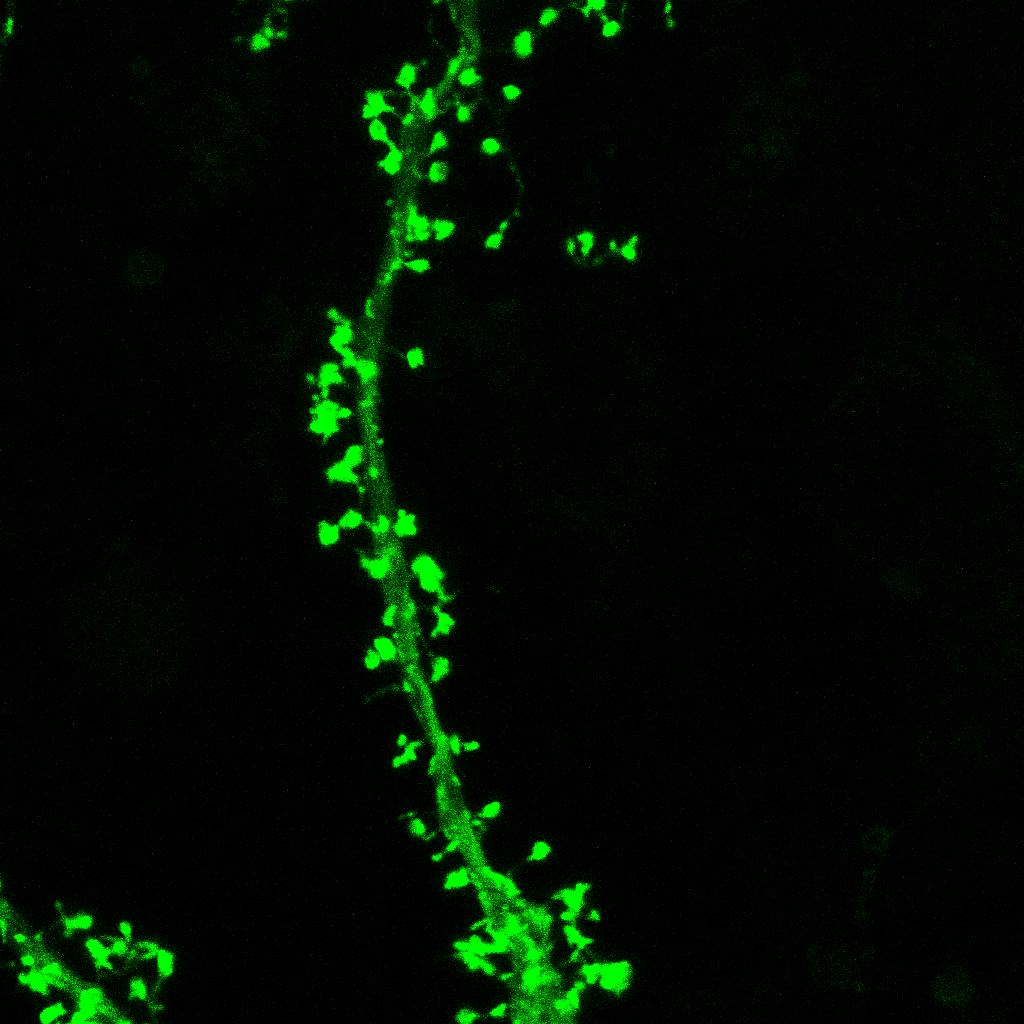}
        }
        \subfigure{%
            \label{maximum_filter_hold1}
            \includegraphics[scale=0.1]{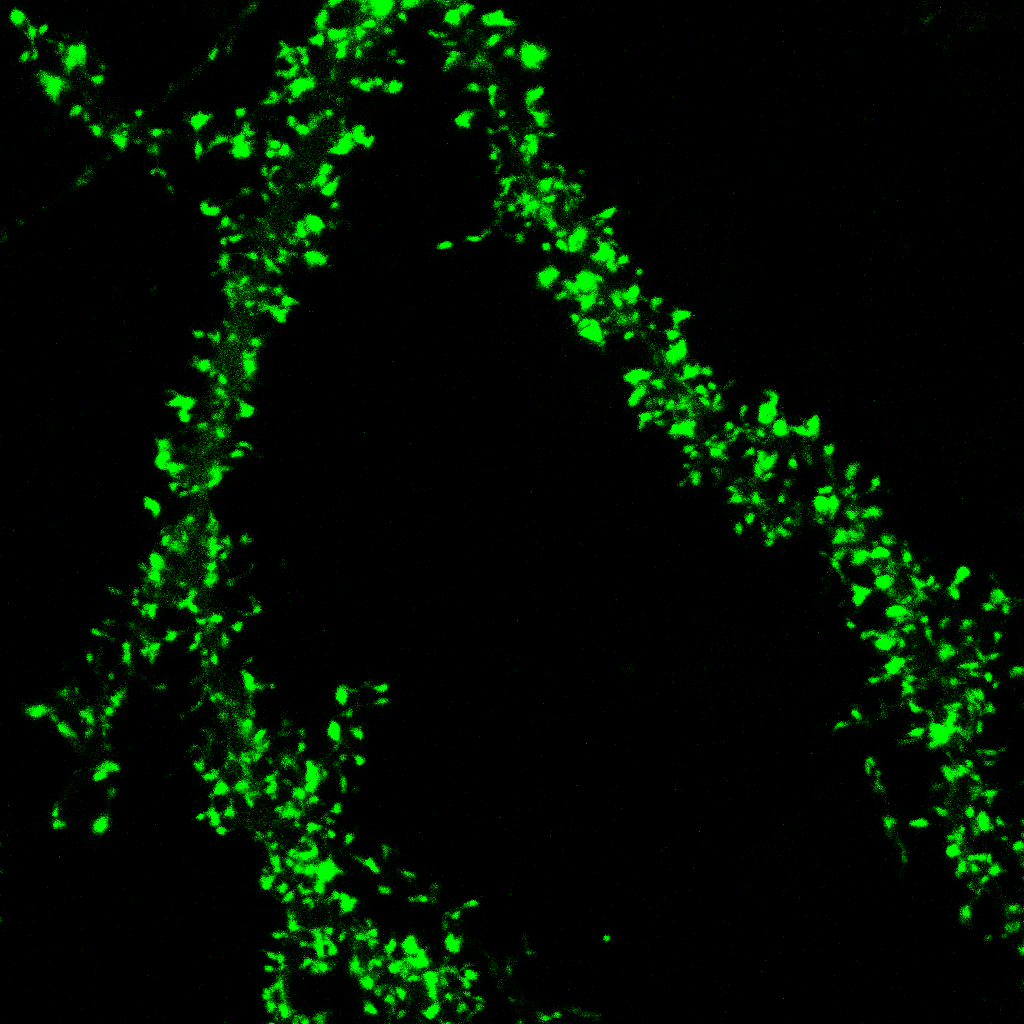}
        }\\
\subfigure{%
            \label{maximum2}
            \includegraphics[scale=0.1]{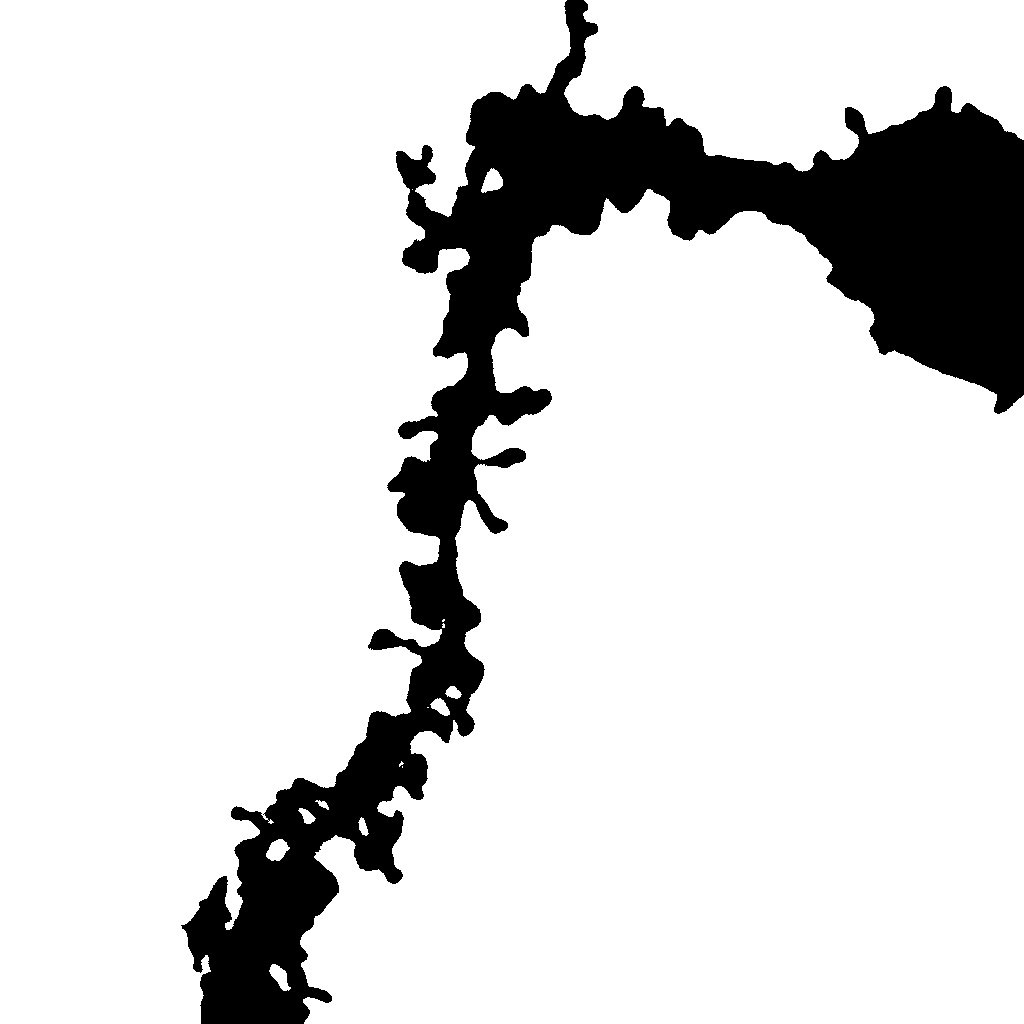}
        }%
        \subfigure{%
           \label{maximum_filter2}
           \includegraphics[scale=0.1]{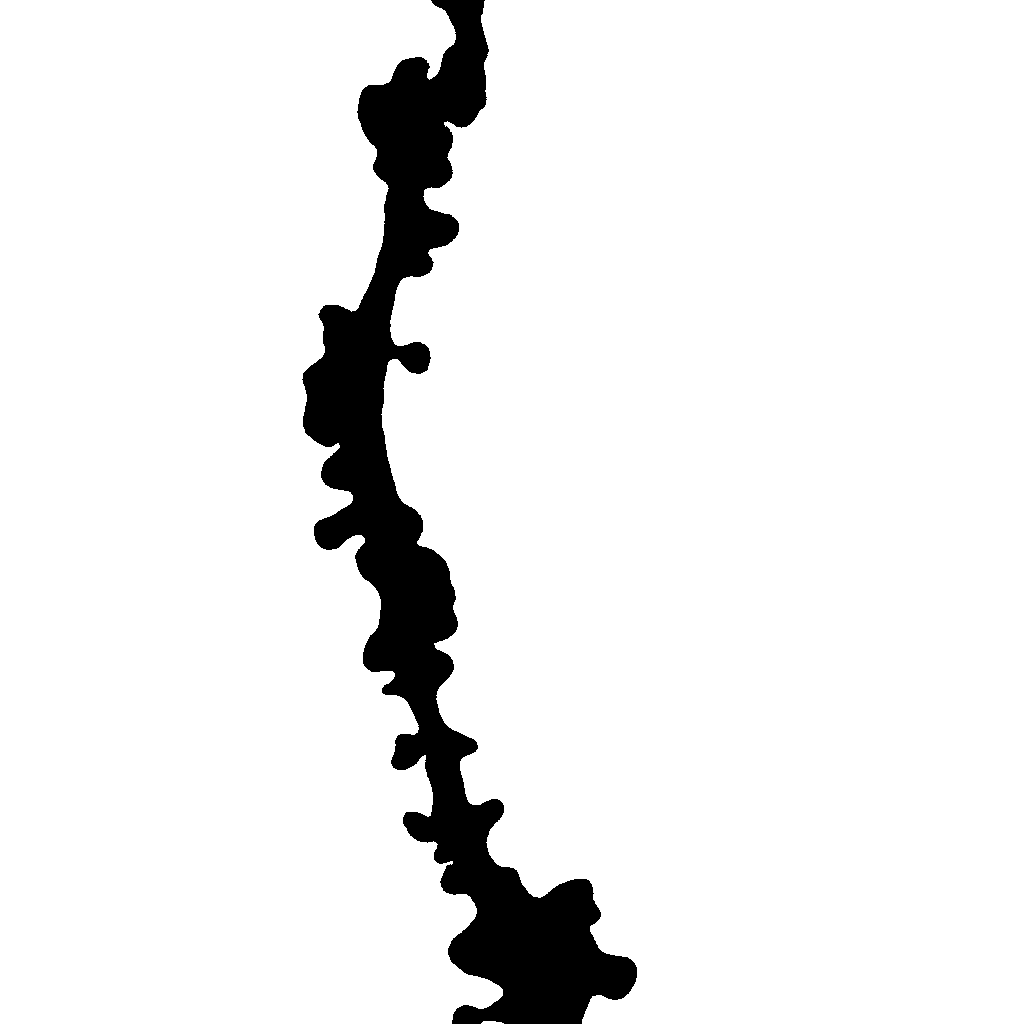}
        }
        \subfigure{%
            \label{maximum_filter_hold2}
            \includegraphics[scale=0.1]{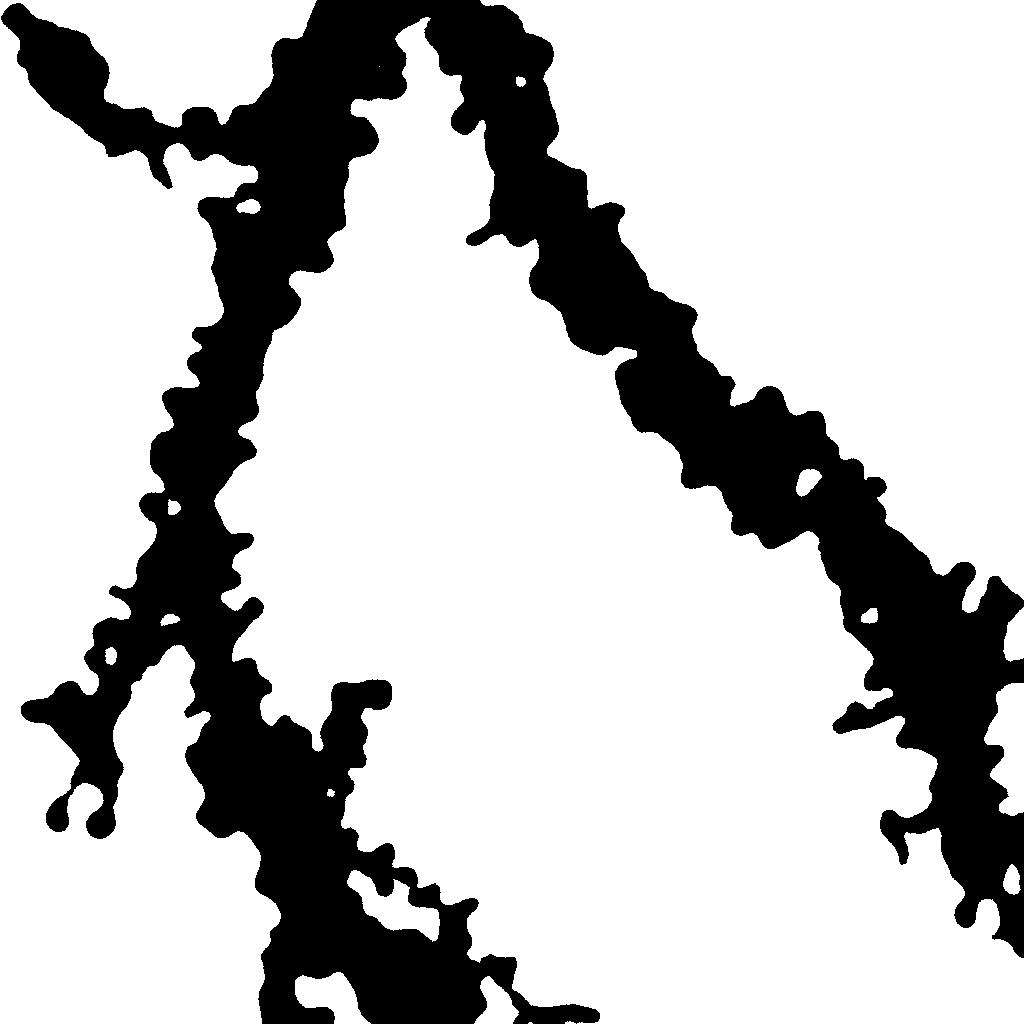}
        }
\caption{Top (from a to c): Three examples of dendritic fragments of hippocampal neurons in culture transfected with Actin-GFP.
Bottom (from a' to c'): Structures obtained with the NeuronpersistentJ application with a median filter 10.}\label{Examples}
\end{figure}

The procedure to detect neural structure presented in the previous section has been implemented
as a new plugin for \emph{ImageJ}~\cite{ImageJ} called \emph{NeuronPersistentJ}~\cite{NeuronPersistentJ}.
Figure~\ref{Examples} illustrates the results which are obtained with NeuronPersistentJ using
three different examples considering $10$ as the length of the median filter. As can be seen in such
examples both the noise and structures of neighbor neurons are removed from the final result.

We have validated our method and plugin with a set of image stacks of real 3D neuron dendrites acquire
using the procedure explained in Section~\ref{sec:bpaia}. In order to test the suitability of our software,
we have compared a manual selection of the region of interest with the results obtained using NeuronPersistentJ.
The manual selection was performed using the polygonal selection tool from ImageJ. In order to compare the
two tracings (the manual and the one obtained using NeuronPersistentJ), we have considered both the \emph{accuracy}
and the \emph{efficiency}.

The accuracy of the plugin is measured with the three following relevant features: $(1)$ the number of branches
obtained with the manual tracing compared with the number of branches detected with NeuronPersistentJ, $(2)$ the
area of the region selected manually recognized with NeuronPersistentJ (that is, the intersection, $\cap$, of the
region selected manually and the one obtained with NeuronPersistentJ) and $(3)$ the area of the region detected by
NeuronPersistentJ which does not appear in the manual tracing with respect to the area which does not contain the
manual tracing (i.e. the area of the region recognized by NeuronPersistentJ minus, $\setminus$, the region manually
selected with respect to the complement, $\_^C$, of the region manually selected). To compute the percentages associated
with these features, we use the following formulas.

\begin{eqnarray}
   (1) & = & \frac{\texttt{Number of branches of NeuronPersistentJ tracing}}{\texttt{Number of branches of manual tracing}} \times 100\nonumber \\
   (2) & = & \frac{\texttt{Area (NeuronPersistentJ tracing} \cap \texttt{Manual tracing)}}{\texttt{Area (Manual tracing)}} \times 100 \nonumber \\
   (3) & = & \frac{\texttt{Area (NeuronPersistentJ tracing} \setminus \texttt{Manual tracing)}}{\texttt{Area((Manual tracing)}^C)} \times 100 \nonumber
\end{eqnarray}

It is worth noting that the higher the values for both $(1)$ and $(2)$ the better since this means that
we are close to detect all the branches and the whole region of interest. On the contrary, the value of $(3)$
should be small in order to avoid the inclusion of regions which are not relevant.

The experimental results that we have obtained with our dataset, considering different lengths for
the median filter, using NeuronPersistentJ are shown in Table~\ref{table}. As we are seeking an
equilibrium between the values of the features $(2)$ and $(3)$, the best value for the length of the
filter is $10$.

\begin{table}
{%
\begin{center}
\begin{tabular}{|c|c|c|c|c|c|c|}
\hline
\backslashbox{length of filter}{percentage} & $(1)$  & $(2)$  & $(3)$ \\
\hline
5  & $96.2\%$ & $78.06\%$  & $4.43\%$  \\
\hline
10  & $98.2\%$ & $93.3\%$  & $4.19\%$  \\
\hline
15  & $98.7\%$ & $94.9\%$  & $6.25\%$  \\
\hline
\end{tabular}
\end{center}
}%
\caption{Percentages of accuracy of NeuronPersistentJ. $(1)$ Percentage of components detects with NeuronPersistentJ
versus manual tracking. $(2)$ Percentage of area detected with NeuronPersistentJ versus manual tracking. $(3)$
Percentage of area draw by NeuronPersintentJ not present in the manual tracking. Percentage is the mean value
from eight images.}\label{table}
\end{table}

Let us consider the efficiency of the plugin. As we have explained previously the manual method to select
the region of interest consists in using the polygonal tool of ImageJ in the maximum projection image.
This manual procedure takes approximately three minutes per neuron. On the contrary, the results are obtained
in half the time using NeuronPersistentJ. This is quite relevant since in order to test the effect of some
experimental treatments over neurons we do not study just one neuron but batteries of neurons. Therefore, the use of
NeuronPersistentJ means a decreasing of the time invested to detect the neuronal structure.

In view of the results, our method can be considered as a good approach, both from the accuracy and
efficiency point of view, to automatically trace neuronal morphology from z-stacks.

\section{Discussion}\label{sec:dis}

The geometric persistence method reported here has been used to develop an ImageJ/Fiji
plugin able to extract the neuronal structure. In particular, the contour of the neuron 
is segmented and therefore the region of interest is recognized. 

The application is based on the fact that the neuron structure is present, ``or persists'',
in all the levels of z-stack images. Our plugin, automatically, generates a digital 2D 
representation of a three-dimensional neuron in the final picture. After the extraction process,
structures from neighbor neurons, background noise and unspecific staining are eliminated from 
the final image.  

The plugin works analyzing every optical plane and comparing the maximum intensity projection with
the slides of the z-stack. After a preprocessing step where the salt-and-paper noise is removed, 
using the median filter, from the slides of the z-stack and the maximum intensity projection, the plugin removes
from the maximum intensity projection the elements which does not live enough (i.e. the elements which do not
appear in all the slices of the stack) obtaining as result the structure of the neuron. 

Histological and imaging protocols are crucial to determine the number of neurons that would be stained and their
relative signal to the background of the images. In this report we have chosen the transfection of a GFP
protein (Actin-GFP) in hippocampal neurons in culture. Electroporation after plating renders a large number
of transfected cells 48 hours after electroporation~\cite{Mor00}. During the successive days in culture neuronal
density of transfected cells decay slowly to a final density of $20-30$ neurons in a $12$ mm coverlips. 
Therefore, this culture conditions allows the growing of fully develop neurons withing a broad distance
from another transfected neuron. The high GFP quantal yield,
results in a excellent contrast staining, improving signal/background ratio. Moreover the plasmid vector employed
a PDGF-neuronal promoter, ensuring physiologic levels of expression; as it was previously reported cells
could undergo Actin-GFP expression without further developmental problems~\cite{Mor00}. 

To validate our method we have compared a manual surface tracking employing the polygonal selection from ImageJ.
The validation uses as a control the total area delimiting by a manual tracing and compares it with the area delimited
by NeuronPersistentJ. It is worth noting that the result of the comparison depends on the value of the length of the low
pass filter selected; large values will led to a broad structure, on the contrary small values will produce sharp and 
more defined images. Employing this validation method our results indicate that NeuronPersistentJ is suitable to carry
out the recognition of the neuron structure. 

The number of manipulations during the reconstruction process is always a drawback
for a fully automatic process. NeuronPersitentJ requires a set of binary images; thus, selection of the length of a 
low-pass filter value that retains the maximal information from the z-stack pictures is the only parameter determined 
by the experimenter and clearly it depends on the images conditions. NeuronPersitentJ, as other public or commercial
available plugins, works better with highly contrasted images, such the ones obtained by inmunofluorescence.
The topological approach employed here and the use of binary pictures is independent from the nature of the picture. 
However, as mentioned, highly contrasted pictures and a clear and continuous staining are key elements for a fine
reconstruction.

Skeletonization of neuronal structure has been a popular solution to neuronal reconstruction and structure 
extraction~\cite{Neurontracing}. Most of the public and commercial solutions available require a manual or semi 
manual process. Drawing the structure, connecting the dendritic fragments or selecting end or branching points 
are typical approximations. NeuronPersitentJ renders an automatically structure into a 2D image that contains 
only the connected dendrites of the neuron. The image can easily been implemented to an automatic Sholl analysis to quantify
changes in dendritic morphology. Even though, our plugin must be consider as a first step in 
the full process of reconstruction.

As mentioned, NeuronPersistentJ would be used a basic towards the automatic detection and clasification of diferent
features of neuronal structure, such spine density or dendritic arborization.

\section*{Information Sharing Statement}

NeuronPersistentJ is available through the ImageJ wiki (\url{http://imagejdocu.tudor.lu/}) 
using the link \url{http://imagejdocu.tudor.lu/doku.php?id=plugin:utilities:neuronpersistentj:start}.
NeuronPersistentJ is open source; it can be redistributed and/or modified under the terms of the GNU 
General public License.



\bibliographystyle{spmpsci}
\bibliography{ndisiaphi}

\end{document}